\documentclass[10pt,twocolumn,letterpaper]{article}

\usepackage{wacv}
\usepackage{times}
\usepackage{epsfig}
\usepackage{graphicx}
\usepackage{amsmath}
\usepackage{amssymb}

\wacvfinalcopy 

\def\wacvPaperID{483} 

\ifwacvfinal\fi
\begin{document}

\def\x{{\mathbf x}}
\def\L{{\cal L}}
\def\vcoco{{V-COCO}}
\def\hibiscus{{CALIPSO}}
\def\triplet{{$<\mathit{subject, verb, target}>$ }}
\def\aprole{{$AP^1_{role}$ }}
\def\aprolebf{{$\mathbf{AP^1_{role}}$ }}	
\def\retina{{RetinaNet}}
\def\r50{{Faster R50}}
\def\resnext{{Faster RNext101}}
\def\pairbased{{\emph{pair-based}}}
\def\pairfree{{\emph{pair-free}}}

\title{Classifying All Interacting Pairs in a Single Shot}

\author{Sanaa Chafik$^{\star\dagger}$ \hspace{1.6cm} Astrid Orcesi$^{\star\dagger}$ \hspace{1.6cm} Romaric Audigier$^{\star\dagger}$ \hspace{1.6cm} Bertrand Luvison$^{\star\dagger}$\\
$^{\star}$ CEA, LIST, Vision and Learning Lab for Scene Analysis, PC 184, F-91191 Gif-sur-Yvette, France\\
$^{\dagger}$ Vision Lab, ThereSIS, Thales SIX GTS, Campus Polytechnique, Palaiseau, France\\
{\tt\small \{firstname.lastname\}@cea.fr}
}

\maketitle
\ifwacvfinal\thispagestyle{empty}\fi

\begin{abstract}
In this paper, we introduce a novel human interaction detection approach, based on \hibiscus{} (Classifying ALl Interacting Pairs in a Single shOt), a classifier of human-object interactions.
This new single-shot interaction classifier estimates interactions simultaneously for all human-object pairs, regardless of their number and class.
State-of-the-art approaches adopt a multi-shot strategy based on a pairwise estimate of interactions for a set of human-object candidate pairs,
which leads to a complexity depending, at least, on the number of interactions or, at most, on the number of candidate pairs.
In contrast, the proposed method estimates the interactions on the whole image.
Indeed, it simultaneously estimates all interactions between all human subjects and object targets by performing a single forward pass throughout the image.
Consequently, it leads to a constant complexity and computation time independent of the number of subjects, objects or interactions in the image.
In detail, interaction classification is achieved on a dense grid of anchors thanks to a joint multi-task network that learns three complementary tasks simultaneously:
(i) prediction of the types of interaction, 
(ii) estimation of the presence of a target and
(iii) learning of an embedding which maps interacting subject and target to a same representation, by using a metric learning strategy.
In addition, we introduce an object-centric passive-voice verb estimation which significantly improves results. 
Evaluations on the two well-known Human-Object Interaction image datasets, \vcoco{} and HICO-DET, demonstrate the competitiveness of the proposed method (2nd place) compared to the state-of-the-art while having constant computation time regardless of the number of objects and interactions in the image.
\end{abstract}

\section{Introduction}
Several tasks of computer vision address the problem of understanding the semantic content of images,
like visual relationship recognition.
More specific than visual relationship, Human-Object Interaction (HOI) detection aims at detecting what happens and where in the image by paying exclusive attention on human-centric interactions.
HOI detection is a challenging problem, essential for various applications such as activity understanding, surveillance, ambient assisted living, cobotics, etc. 
In the case of surveillance system, quickly understanding human-centric interactions is particularly interesting. As images may contain possibly numerous people and interactions, it is crucial for an HOI detection method to be scalable with the number of visible objects and interactions.
This scalability issue motivated our work. 
In the following, ``objects'' assigned with human class are called \emph{subjects} while those with non-human class are \emph{targets}.
More formally, HOI detection consists in determining and locating the list of triplets \triplet describing all the interactions visible in the image.
Although HOI detection was classically based on video 
(in general, with a focus on a single action), recent approaches based on a single image have shown impressive results on detecting simultaneous interactions.

Generally speaking, image-based HOI detection task is achieved by solving the following sub-tasks: detecting interacting objects (the \emph{object detection problem}), correctly pairing such objects (the \emph{association problem}) and classifying the interactions (the \emph{verb classification problem}).
Most approaches~\cite{HORCNN, ICAN, InteractNet, VSRL, Transferable, GPNN, Rowan, GraphContrastive} rely on an object detector that identifies some candidates for subject-target pairs which boxes are then processed in a second step to assess interaction presence and type. Sometimes, features for objects and pairs are first extracted and, then, processed to infer object class, location, subject-target association and verb classification~\cite{AssEmbd}.
Thus, all these methods have a pair-based second-step processing, which may become a scalability issue when dealing with large numbers of object and interaction instances in the image.

This work proposes a new interaction detection approach, named \hibiscus{} (Classifying ALl Interacting Pairs in a Single shOt) which complexity is independent of the number of interactions.
The proposed model simultaneously estimates all interactions between all objects with a single forward pass throughout the image.
It manages the problems of association and verb classification while any external object detector can be used to deal with the problem of object detection.
To this end, \hibiscus{} approach exploits a multi-task learning scheme, performing three complementary tasks: a classification task predicts the verb of interaction, a target presence estimation task assesses the presence of the target object of the interaction and an embedding task maps a pair of interacting subject-target to a 
similar representation.
Lastly, at inference time, any object detector can be used to point out objects of interest and output the corresponding interactions.
Notice that the proposed approach does not use any ontology information such as a prior list of interactions of interest, in order to promote generalization over target classes.
We have evaluated the efficiency of the proposed approach on two widely used HOI datasets. Our results compare favorably (2nd place) with state-of-the-art approaches while having constant computation time regardless of the number of objects and interactions in the image.

\section{Related Work}
\label{sec:relatedwork}

\textbf{HOI \& visual relationships} 
Despite the rapid research progress in analysis of humans and their activities by computer vision, human interaction recognition from a single image remains a challenge. Whereas videos contain rich temporal clues, such as those used in interaction analysis of egocentric videos \cite{Baradel, NextActive, LSTA}, images contain a lot of contextual information that is meaningful to infer relationships between objects.
One of the main problems of detecting visual relationships is the need for tremendous amounts of varied examples, 
as appearances and classes of both subject and target should vary for generalization of each interaction class.
The release of large datasets \cite{HORCNN, VSRL, VG, zhuang2017care} has allowed the developement of several visual relationship detectors in recent years \cite{VSR1, BARCNN, VIPCNN, VSR2, AssEmbd, Ruichi, Rowan, VSR3, GraphContrastive} as well as HOI detectors \cite{Functional, HORCNN, ICAN, InteractNet, VSRL, analogies, GPNN, Knowledge}.

Gupta et al. \cite{VSRL} successively detect a subject, classify the action and associate the target according to an interaction score. 
Several approaches \cite{HORCNN, ICAN, InteractNet, Transferable} extend an object detector model, namely Faster R-CNN \cite{FasterRCNN}, with extra branches either for predicting actions, estimating a probability density over the target object location for each action \cite{InteractNet}, the spatial relations of human-object pairs \cite{HORCNN}, an instance-centric attention measure \cite{ICAN}, or filtering non-interactive human-object pairs with 
cross learning datasets \cite{Transferable}. 
Qi et al. \cite{GPNN} present a generic framework combining graphical models and deep neural network, capturing human-object interactions iteratively.
Li et al. \cite{VIPCNN} introduce a cross branch communication with phrase-guided message to ensure a joint modeling of action classification and target association.

Some techniques~\cite{Functional, Knowledge, Ruichi}
incorporate \emph{linguistic knowledge} to address the issue of having a long-tail distribution of human-object interaction classes.
They exploit the contextual information present in the language priors learnt with a `word2vec' network, to generalize interactions across functionally similar objects.
Alternatively, Peyre et al. \cite{analogies} learn a visual relation representation combining compositional representation for subject, target and predicate with a visual phrase representation for HOI detection. 
Unlike these approaches, our method does not use additional linguistic data.

However, all these approaches have a pair-based processing step, i.e., a substantial processing applied on a set of subject-target pair proposals. This may become a scalability issue when dealing with large numbers of object and interaction instances in the image.
In contrast, we propose a new interaction detection approach which complexity is independent of the number of interactions in the image. The model classifies interactivity on a dense sampling of all possible object locations simultaneously.

\textbf{Metric learning}
has been applied to many different tasks, from image retrieval~\cite{Frome07}, to face recognition~\cite{FaceNet}. In addition to providing a similarity measure to compare images, it can also been used to map visual and text features to a shared feature space~\cite{Frome13, Karpathy15} or associate features of visual elements to recognize a group of such elements. For example, Newell and Deng proposed associative embedddings to group together body joints for human pose estimation~\cite{Newell16}. 
Metric learning is also applied to visual relationship detection~\cite{AssEmbd, analogies, GraphContrastive}.
In particular, Pixel2Graphs~\cite{AssEmbd} produces in a single-shot manner a set of objects and interaction links represented by a graph which is deduced from two heatmaps. Then, in a second step, each of these object or connection features is passed through a fully connected network to predict interaction properties (verb, subject-target association, object class and bounding box). This second step is, thus, dependent on the number of interactions.
Besides, when multiple relations are grounded in the same location, a fixed number of slots are used to manage these overlapped relations, which may be limiting for densely populated images.

\hibiscus{} is also based on the metric learning paradigm. But, in contrast to Pixel2Graphs, it does not use graph to explicitly model each object and each relation. 
Rather, it simultaneously provides associative features and interaction types for all locations of potential subjects and targets in a single shot.
Another fundamental difference is that Pixel2Graphs aims to define a unique feature for each object regardless of the relation verb, and a unique feature for each relation.
Differently, \hibiscus{} aims to define, for each interaction verb, an embedding where all objects involved in an interaction instance should have similar features. 
This allows a subject-target pair to have multiple interactions while solving the overlapped interaction issue. Moreover, having a different embedding space for each verb should intuitively leave more flexibility for modeling very different types of interactions (contact interaction, distant interaction, etc.).

\section{Proposed Method}
\label{sec:proposedmethod}
In this section, we present our proposed approach, named \hibiscus{}, for interaction modeling. The task of human-object interaction detection consists of locating and recognizing humans and objects in a given image and identifying the actions (i.e. verbs) that connect them. Formally, locating and recognizing the set $\mathcal{T}$ of interaction triplets \triplet with $verb$ an interaction verb among $V$ verbs. The proposed approach deals with associating and classifying subject-target interacting pairs 
with complexity independent of the number of interactions.
To this end, \hibiscus{} decorrelates object detection task from the association and the interaction classification tasks. It requires an object detector only at inference time, in order to point out and classify the objects to be really considered for interaction.
We first give an overview of the proposed approach, then detail the proposed model tasks. Last, we describe the inference process.
\subsection{Overview}
\begin{figure*}[h!]
	\centering
	\includegraphics[width=12.5cm]
	{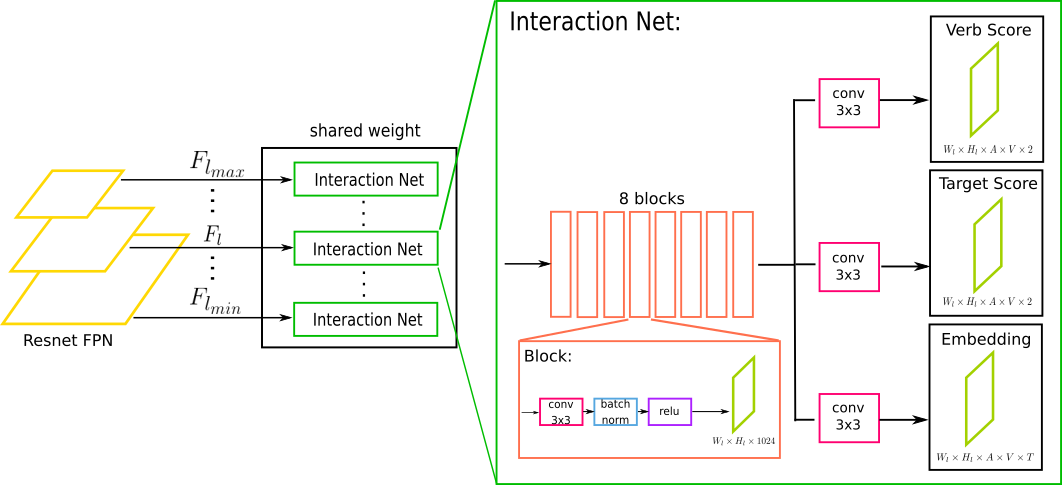}
	\caption{\hibiscus{} architecture starts with a Resnet FPN backbone with feature pyramid ($F_{l_{min}}$ to $F_{l_{max}}$). The feature of a given level $l$ has size $W_l \times H_l$. The interaction network is applied on each level. It is composed of a succession of 8 blocks. The network splits into 3 branches computing 3 complementary tasks. $A$ is the number of anchors, $V$ is the number of verbs and $T$ is the size of the embedding.}
	\label{fig:overview}
\end{figure*}
The proposed model architecture is a multi-task neural network (cf. Figure~\ref{fig:overview}). It consists of a backbone network and an interaction network. From an image $I$, a feature pyramid is constructed using a Feature Pyramid Network (FPN) \cite{FPN} backbone, capturing multi-scale high-level semantics. The FPN backbone network takes an input image $I$ of size $W\times H$, and outputs multiple level feature maps $F_{l}$ of size $W_l \times H_l$, where $W_l= \frac{W}{2^{l}}$, $H_l = \frac{H}{2^{l}}$ and $l$ is the pyramid level, $l\in [l_{min}, l_{max}]$. The FPN is built on top of the Residual Network following RetinaNet \cite{RetinaNet} architecture.

Then, the featurized image from each FPN level feeds a fully convolutional interaction network ending with three tasks. 
The first task is an action classification that predicts the verb describing the type of interaction between subject and target. 
The second task is a target presence estimator providing the probability that the object, a human is interacting with, is visible or not, for a given verb. 
The third task associates interacting subject and target, by mapping them to the same representation. The overall network is trained end-to-end, the three tasks are trained simultaneously, sharing the common backbone which subsequently helps generalization by regularizing training.
\hibiscus{} approach simultaneously estimates all possible interactions between all humans and objects in the image, with a single forward pass through the architecture. Thus, \hibiscus{} is independent of the number of subjects, targets and interaction instances.  
Moreover, by densely estimating embeddings for each verb, negative example mining is exhaustive over the image. For example, all people not doing a specific action over the image will be provided to the network as negative samples to learn the embedding space of this action.

Finally, at inference, after generating dense maps, an external object detector is used to point out candidate subjects and targets. Therefore, the final interaction triplets are determined thanks to the object class of targets provided by the detector together with the association information and interaction verb given by \hibiscus{}.

\subsection{Interaction module}

Firstly, interaction detection requires to identify the (human and non-human) objects in interaction. At each feature map location, a set of reference boxes called anchors are defined. These anchors are of multiple scales and aspect ratios aligned to objects. We use anchor boxes similar to those in the Region Proposal Network of \cite{RCNN}. For each level $l$ of the feature pyramid, we define a set of anchors $\mathcal{A}_{l}$, containing $W_l\times H_l\times A$ anchors, where $A=9$ is the number of anchors at each feature map location. For sake of clarity, we define $\mathcal{A}=\left \{ a_{i} \right | i \in [1,A_{all}]\}$ as the set of all the anchors over the pyramid, where $A_{all}$ is the total number of anchors. Each anchor in $\mathcal{A}$ is labeled as foreground or background. We denote $\mathcal{G}=\left \{ g_{j} \right | j \in [1,B]\}$ as the set of ground-truth bounding boxes, where $B$ is the number of objects in the image. As it is classically done \cite{RetinaNet}, an anchor is assigned to a ground-truth box if its intersection-over-union (IoU) is over $0.5$. We define $\mathcal{A}_{g_{j}}$ as the set of anchors assigned to ground-truth boxes $g_{j}$ and $\mathcal{A}_\mathcal{G}$ the union of all the anchors assigned to a ground-truth box.

The interaction subnet is responsible for three tasks learnt simultaneously. This subnet is applied with the same weights to each level of the backbone feature pyramid, capturing the relationships between instances of different sizes that occurred in different levels of the FPN. Moreover, the shared weights of the network applied to each pyramid level enhance the learning of correlated tasks. 
These tasks share a succession of eight blocks of convolution, batchnorm and ReLU layers. The number of blocks was found empirically (cf. section \ref{section:ablation}). The spatial size of each task output for a given pyramid level $l$ is equal to the feature map size at this level: $W_l\times H_l$.
\\
\\
\textbf{Verb prediction task:}
Considering that the subjects of the interaction can take simultaneously multiple actions, the verb prediction task minimizes a multi-label binary cross entropy loss $L^{verb}$ between the predicted and the ground-truth verbs. 
Unlike other methods, we introduce an additional object-centric passive verb estimation to reciprocally improve the relationship detection. 
The verb prediction task is performed based on the contextual appearance which is very informative to distinguish actions that humans carry out and objects undergo. 
Among $\mathcal{A}_\mathcal{G}$, we find the active anchors, representing anchors associated to people executing the actions, and the passive anchors associated to objects undergoing the action. 
The passive form classification is an optional task that improves performances (cf. Section~\ref{section:ablation}).
Verb prediction task then produces, for each anchor, a classification output over the verbs in both active and passive forms resulting in an output of size $2V$ with $V$ the number of different verbs.
This reciprocal interaction estimation is expected to be a soft way to enforce the interaction verb classification by merging human-centric and object-centric information.
\\
\\
\textbf{Target presence estimation task}
is a complementary task to the verb prediction task. It aims to estimate the probability that the object, a person is interacting with, is visible or not.
Similar to the verb prediction task, the target object estimation is performed on the contextual appearance of each person anchor, capturing the spatial position and the surroundings of the person in the image. 
For each anchor, the output of size $2V$ consists of binary sigmoid classifiers. The objective of the training is to minimize the binary cross entropy loss, $L^{target}$, between the ground-truth target object labels and predicted target estimation.
\\
\\
\textbf{The interaction embedding task}
aims to map several anchors corresponding to interacting subject and target to the same representation for a given verb.
The embedding subnet is a function mapping the anchor space $\mathcal{A}$ to a new space such that: 
$emb \colon \mathcal{A} \to \mathbb{R}^{V \times T}$ 
where $T$ is the dimension of the interaction embedding space specific to one verb.
For a given verb, this embedding task aims at ensuring to assign, first, the same embedding to anchors related to the same object instance and, second, the same embedding to anchors belonging to the same interaction. 

Formally, given anchors $a_{i}, a_{j} \in \mathcal{A}_\mathcal{G}^2$, $a_{i}$ and $a_{j}$ are interacting according to verb $v$, i.e. $a_{i}\sim_{v} a_{j}$, if:

\begin{equation}
\label{eq:sim_def_intra}
\exists\ g_n\in \mathcal{G} \ | \ (a_{i}, a_{j}) \in \mathcal{A}_{g_{n}}^2\\ 
\end{equation}

or

\begin{equation}
\label{eq:sim_def_inter}
\left.
\begin{matrix}
\exists \ {(g_n, g_m) \in \mathcal{G}^2}, \ n\neq m, \\
<g_{n}, v, g_{m}>\text{or} <g_{m}, v, g_{n}> \in \mathcal{T}
\end{matrix}
\right| 
(a_{i},a_{j}) \in \mathcal{A}_{g_{n}} \times \mathcal{A}_{g_{m}}
\end{equation}

Accordingly, to each verb $v$, corresponds a set of equivalence classes associated with an equivalence relation $\sim_{v}$, denoted by $\mathcal{C}_{v} = \{ \mathcal{C}_{v}^{i} | i\in [1,E_v] \}$, with $E_v$ the number of equivalence classes for verb $v$. 
Let $|\mathcal{C}_{v}^{i}|$ be the number of anchors belonging to the equivalence class $\mathcal{C}_{v}^{i}$. 
The reference of the equivalence class is defined by the mean of the output embeddings of the same equivalence class as follows:
\begin{equation}
\overline{e_{\mathcal{C}_{v}^{i}}}=\frac{1}{|\mathcal{C}_{v}^{i}|}\sum_{j \in \mathcal{C}_{v}^{i} }e_{j}^{v}
\end{equation}
where $e_{j}^{v}$ is the predicted embedding for the anchor $a_j$ and verb $v$.

The embedding network aims to learn the equivalence class space $\mathcal{C}_{v}$, by minimizing the equivalence loss $\mathit{L_{v}^{emb}}$, defined in a metric learning form. For a given verb $v$, the loss is defined as:
\begin{equation}
\mathit{L_{v}^{emb}} = \mathit{L_{v}^{pull}} + \mathit{L_{v}^{push}}  
\end{equation}

The pulling loss that aims at gathering the corresponding elements, is defined as:
\begin{equation}
\mathit{L_{v}^{pull}} = \frac{1}{E_{v}}
\sum_{\mathcal{C}_{v}^{i} \in \mathcal{C}_{v}}\frac{\lambda_{\mathcal{C}_{v}^{i}}}{\left |  \mathcal{C}_{v}^{i}\right |}
\sum_{j \in \mathcal{C}_{v}^{i}}\left ( e_{j}^{v}-\overline{e_{C_{v}^{i}}} \right )^{2}\\
\end{equation}

Based on the ground truth annotations defining interacting instances, the first term of the equation aims to merge interacting instances to the same equivalence class by computing the mean squared distance between the equivalence references $\overline{e_{C_{v}^{i}}}$ and the predicted embedding $e_{j}^{v}$ for each anchor $j$ in equivalence class $\mathcal{C}_{v}^{i}$.
The weight $\lambda_{\mathcal{C}_{v}^{i}}$ aims at focusing more on equivalence classes representing real interacting subjects and targets rather than equivalence class associated to a single object not belonging to any interaction (cf. equation \ref{eq:sim_def_intra}). It is defined as:

\begin{equation}
\lambda_{\mathcal{C}_{v}^{i}} = 
\begin{cases}
\lambda_{pull} & {\text{if} \ \exists \ a_j,a_k \in \mathcal{C}_{v}^{i} \ \text{such that}} \\
& (a_{j},a_{k}) \in \mathcal{A}_{g_{n}} \times \mathcal{A}_{g_{m}}, n\neq m, \\
& <g_{n}, v, g_{m}> \text{or} <g_{m}, v, g_{n}> \in \mathcal{T}; \\
1			   & \text{otherwise.}
\end{cases}
\end{equation}

The pushing loss enables the mapping of not interacting instance anchors into different clusters using an exponential decreasing function with fixed parameter $\sigma$. It is defined as:
\begin{equation}
\mathit{L_{v}^{push}} = \frac{1}{E_{v}^2}
\sum_{\underset{i\neq j}{\mathcal{C}_{v}^{i}, \mathcal{C}_{v}^{j} \in C_{v}^2}} \gamma_{\mathcal{C}_{v}^{i}, \mathcal{C}_{v}^{j}} \ exp\left (\frac{-1}{2\sigma ^{2}}\left (\overline{e_{\mathcal{C}_{v}^{i}}}  - \overline{e_{\mathcal{C}_{v}^{j}}}\right )^{2}\right )
\end{equation}

The weight $\gamma_{\mathcal{C}_{v}^{i}, \mathcal{C}_{v}^{j}}$ introduces a soft penalty to the loss to force the network to associate the correct target among several objects present in the image that are usual target for this verb. 
For example, the feature of a person sitting on a given chair should not be clustered with features of other chairs or objects one can sit on (e.g. couch, bed, table, ...), present in the image. 
This pushing weight is a way to enforce the selection of the right target among various candidates even if they are suitable for this interaction.
More formally, let $lab_{i}$ be the class label of anchor $a_{i}$ and $\mathcal{L}_v$ the set of object classes that can be involved in the type of interaction given by verb $v$ according to statistics on the dataset (e.g. chair, couch, bed, table, ... for verb `̀`sit''). The weight $\gamma_{\mathcal{C}_{v}^{i}, \mathcal{C}_{v}^{j}}$ is defined as :

\begin{equation}
\gamma_{\mathcal{C}_{v}^{i}, \mathcal{C}_{v}^{j}} =
\begin{cases}
\gamma_{push} & \text{if} \ \exists (a_k,a_l) \in \mathcal{C}_{v}^{i} \times \mathcal{C}_{v}^{j} \ \text{such that}\\ 
& (a_{k},a_{l}) \in \mathcal{A}_{g_{n}} \times \mathcal{A}_{g_{m}}, \ n\neq m, \\
& \ (lab_k, lab_l) \in \mathcal{L}_v^2; \\
1			  & \text{otherwise.}
\end{cases}
\end{equation}

This embedding scheme is performed for each verb, allowing the network to learn the different ways of interaction depending on the verb. 
Moreover, the embedding predictions are performed simultaneously on all anchors, regardless of the number of object instances. 
This also enables a better management of negative interactions at training by processing all non-interactions in the image. 
In addition, it allows a fast and accurate instance connection at inference. 
Notice that the embedding task does not make specific assumptions between subject and target positions and can thus model both distant and close interactions.
In addition, the embedding task learns to associate objects of possibly different sizes, i.e., localized on different pyramid levels.

The overall loss $L_{total}$ of the proposed model is the sum of verb classification loss $L^{verb}$, the target presence loss $L^{target}$, and the mean of embedding losses $L_{v}^{emb}$. 
\begin{equation}
L_{total}= L^{verb} + L^{target} + \frac{1}{\left | V \right |} \sum_{v \in V} L_{v}^{emb}
\end{equation}

\subsection{Inference}

In the same manner as existing approaches, we predict the HOI triplets \triplet, which involves predicting the human-object bounding box pairs, identifying the verb and the triplet score. 
The three tasks of the proposed model provide three feature anchor maps. 
The feature anchor map of the first task defines the action score of each location in the image. 
The second task provides a feature map estimating for each verb, the presence of an interacting target for each human anchor. 
The third feature anchor map gives an embedding for each anchor in the image, to determine the interacting anchors. 
The method extracts all the feature maps simultaneously and independently of the number of object instances which are at arbitrary image locations and scales, contrary to most existing approaches where every selected human-object pair is processed individually. 

The prediction of HOI triplets requires preliminary to identify all human-object bounding boxes. 
For that purpose, \hibiscus{} requires at inference an external detector to point out anchors of interest from the three feature maps. 
The external detector can be any bounding box-based object detector providing the bounding box positions and the class scores, noted $s_{h}^{det}$ for human and $s_{o}^{det}$ for object. 
The detector provides a set of candidate object bounding boxes that are subsequently mapped to the anchor grid. Hence, from this mapping, for each verb $v$ and for each candidate bounding box, different scores can be read: 
verb scores (specifically, active score $s_{v,h}^{active}$ for human and passive score $s_{v, o}^{passive}$ for object), target presence scores $s_{v,h}^{target}$ for human, and embeddings $e_{i}^{v}$ of each detected instance. These embeddings are compared each other defining a connection score $s_{v,h,o}^{emb}$ computed as follows:
\begin{equation}
s_{v,h,o}^{emb} = exp^{(-|e_{h}^{v} - e_{o}^{v}|)}
\end{equation}

All the above scores together define the triplet score as:
\begin{equation}
s_{v,h,o}^{triplet} = \sqrt[6]{s_{h}^{det} \ s_{v,h}^{active} \ s_{o}^{det} \  s_{v,o}^{passive} \ s_{v,h}^{target} \ s_{v,h,o}^{emb}} 
\label{score_triplet_target}
\end{equation}

All the possible triplets are computed for each detected human and each verb. Additionally, a pair score is computed for target absence case:
\begin{equation}
s_{v,h}^{pair} = \sqrt[3]{s_{h}^{det} \ s_{v,h}^{active} \ (1-s_{v,h}^{target})}
\label{score_triplet_notarget}
\end{equation}

For a given verb and a given person, all triplets and the pair are sorted according to their scores and the one with the highest score is kept after thresholding.

\section{Experiments}
\label{sec:experiments}

Experiments are conducted on two widely used datasets for interaction detection with a comparison between the proposed approach and recent state-of-the-art.

\subsection{Datasets}
\textbf{\vcoco{} dataset}\footnote{https://github.com/s-gupta/v-coco}~\cite{VSRL} is a subset of the COCO dataset~\cite{COCO} for human-object interaction detection. It includes $10,346$ images ($2,533$ images in the train set, $2,867$ images in the validation set and $4,946$ images in the test set). \vcoco{} contains $16,199$ human instances, where each person has annotations for $29$ action categories over $80$ object categories.  The target objects of the dataset are classified into two types: ``object'' or ``instrument'': ``object'' target if it undergoes the action (e.g., ``to cut a cake''), or  ``instrument'' if it is a means enabling the interaction (e.g., ``to cut with a knife''). 
Four verbs do not have target (``stand'', ``smile'', ``run'', ``walk'')

\textbf{HICO-DET dataset} \cite{HORCNN} is a subset of the HICO dataset for human-object interaction detection. It is larger and more diverse than \vcoco{} dataset. HICO-DET includes $47,051 $ images ($37,536$ images in the train set and $9,515$ images in the test set).  HICO-DET contains $117$ action categories over $80$ object categories as COCO dataset. Not all combinations of actions over objects are relevant, according to a defined ontology. As a consequence, only $600$ specific human-object interaction categories are annotated and evaluated.

\subsection{Evaluation metrics}
Following the standard evaluation settings of \vcoco{} \cite{VSRL} and HICO-DET \cite{HORCNN} datasets, we evaluate HOI detection performance using the average precision metrics. The predicted \triplet triplet is considered as a true positive, when all the triplet predicted components are correct. The predicted human and object bounding boxes are supposed to be correct if they have IoU greater than 0.5 with ground truth boxes.

Following previous work \cite{HORCNN, ICAN, InteractNet, GPNN}, the evaluation on \vcoco{} dataset is based on the role mean average precision called \aprole on $24$ verb categories.  Indeed, for the purpose of fair comparison with state-of-the-art approaches, $5$ actions (run, smile, stand, walk and point) are ignored in the evaluation, as done in previous approaches. 

Concerning HICO-DET dataset \cite{HORCNN}, we report the mean AP over three different HOI category sets: (a) all $600$ HOI categories in HICO (\emph{Full}), (b) $138$ HOI categories with less than $10$ training instances (\emph{Rare}), and (c) $462$ HOI categories with $10$ or more training instances (\emph{Non-Rare}).

\subsection{Implementation details}
We initialize the FPN ResNet backbone with corresponding weights of RetinaNet~\cite{RetinaNet} especially trained on COCO dataset from which \vcoco{} images were previously removed.
The \hibiscus{} is trained with stochastic gradient descent (SGD), with an initial learning rate of $0.016$, which is then reduced by $10$ at $25000$ iterations over a batch of size $10$. A horizontal image flipping is applied for data augmentation. The weight decay is set to $10^{-4}$ and the momentum to $0.9$. $\sigma$, $\lambda_{v}$ and $\gamma_{v}$ are experimentally set to $2$, $10$ and $100$.

At inference, \hibiscus{} requires an external detector to filter interacting bounding boxes from the three sub-task feature maps. As done in most state-of-the art methods, the Faster RCNN \cite{FasterRCNN} from Detectron\footnote{https://github.com/facebookresearch/Detectron} framework is used as external detector. It is based on a ResNet-50-FPN backbone to generate all object bounding boxes. Other object detectors are tested to show the influence on HOI detection.

\subsection{Qualitative results}
\begin{figure}[ht!]
	\centering
	\includegraphics[width=\columnwidth]{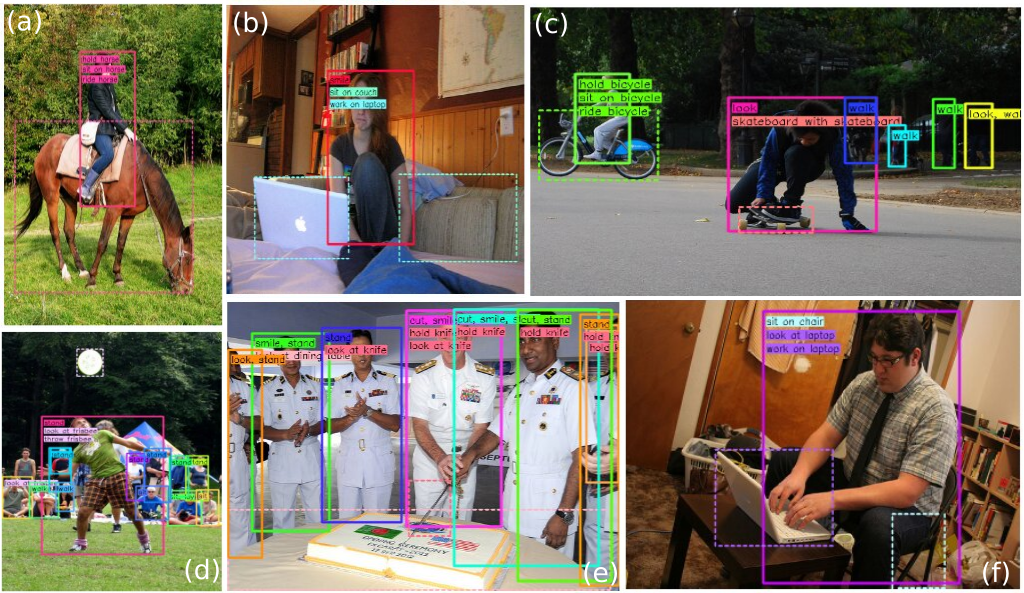}
	\caption{Samples of human-object interactions detected by \hibiscus{} on some \vcoco{} test images. An interaction triplet is composed of a human subject represented by a solid-line box, a target object represented by a dashed-line box and, at the top left of the subject box, the action performed is written on a background with same color as the target object box (best viewed in color).}
	\label{fig:results}
\end{figure}

\begin{figure}[h!]
	\centering
	\includegraphics[width=8.3cm]{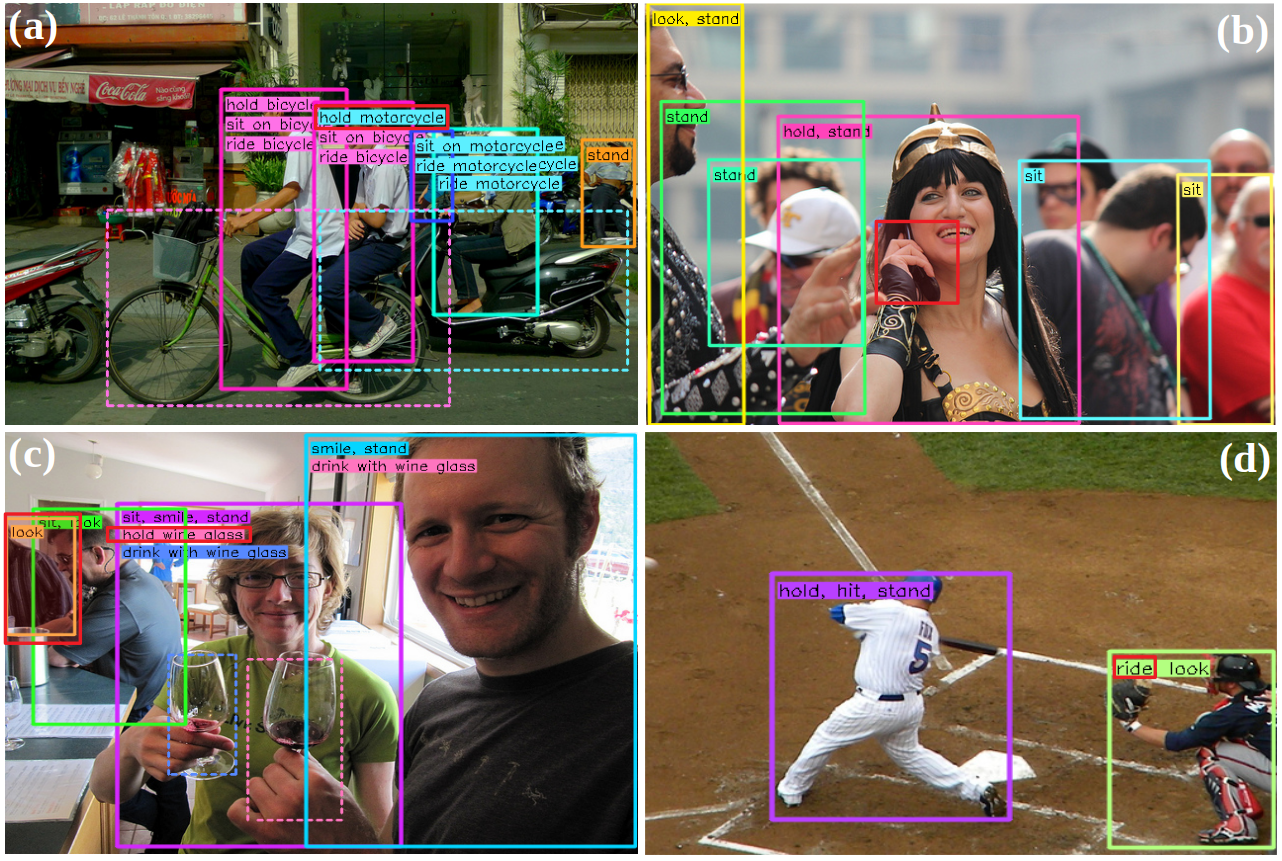}
	\caption{Illustration of some incorrect human-object interaction detections on some \vcoco{} test images.}
	\label{fig:vcoco_fp}
\end{figure}

Figures~\ref{fig:results} and \ref{fig:vcoco_fp} illustrate the interaction results detected by the proposed model. They show all the triplets occurred in the image. Each triplet is represented by a solid-line box for the subject and a dashed-line box for the target object. At the top left of the subject box, the action performed is indicated on a background with same color as the related target box.

Figure~\ref{fig:results} depicts interactions detected by our approach. As can be seen, \hibiscus{} can infer HOI in various situations such as: 
1) Individual person performing different actions on a single object (i.e, ``a person rides, sits and holds a bicycle'', in Figure \ref{fig:results}-a-b-c-f). 
2) Individual person interacting with different objects (e.g., in Figures \ref{fig:results}-b and \ref{fig:results}-f, ``a person works on a computer while sitting on a chair/couch'').
3) Several people interacting with a single object (e.g., in Figure \ref{fig:results}-e, ``two people hold the same knife''). Notice that \hibiscus{} correctly assigns the target object to the corresponding action, and can successfully detect contactless interactions (in Figure~\ref{fig:results}-d, ``look at and throw a frisbee'').

Figure~\ref{fig:vcoco_fp} illustrates another sample of \vcoco{} test images, where \hibiscus{} detects some incorrect triplets. This is mainly caused by : 
1) Wrong object detection, with either no detected object (as shown in Figure~\ref{fig:vcoco_fp}-b where the cell phone is not detected) or misclassified object (illustrated in Figure~\ref{fig:vcoco_fp}-c where the backpack is classified as human). 
2) Wrong verb estimation, depicted in Figure~\ref{fig:vcoco_fp}-d where the person has a confusing posture. 
3)  Wrong target association, shown in Figure~\ref{fig:vcoco_fp}-c where the wine glass is held by the wrong person.
Figure~\ref{fig:vcoco_fp}-a shows an example where all these difficulties appear simultaneously. Indeed, the high density of objects leads to more occlusion, misunderstanding of the object depth in the scene and, thus, confusions in subject-target associations.

\subsection{Quantitative results}

\subsubsection{Ablation study}
\label{section:ablation}

In Table \ref{table:vcoco_abla} we evaluate on \vcoco{} dataset the contributions of various components of the method.

\begin{table}[h!]
	\begin{center}
		\begin{tabular}{| l | l |}
			\hline
			\textbf{Method} & \aprolebf (\%)\\
			\hline
			\textbf{\hibiscus{}} & \textbf{46.36}\\			
			\hibiscus{} w/o weight sharing & 43.86\\
			\hibiscus{} w/o passive mode & 36.86\\
			\hibiscus{} w/o target presence\hspace{2cm} & 25.51\\
			\hline
			\hline
			\hibiscus{} 5 blocks & 44.35\\
			\textbf{\hibiscus{} 8 blocks} & \textbf{46.36}\\
			\hibiscus{} 11 blocks & 45.05\\
			\hline
		\end{tabular}
	\end{center}
	\caption{Ablation studies for \hibiscus{} on the\vcoco{} test set.
	}
	\label{table:vcoco_abla}
\end{table}

\textbf{Shared weights:} Sharing weights across feature pyramid network levels shows improvement by $2.25$ p.p. (percentage points) in interaction detection performances. Intuitively, it may capture better the relationships between instances belonging to different levels of the FPN corresponding to different object size.

\textbf{Passive mode:} Whereas active mode is subject-centric, the passive mode is a way of introducing a complementary target-centric point of view and, thus, introducing redundancy to improve robustness. Without passive mode task, our model reaches an \aprole of $36.86\%$. It increases by approximately almost $10$ p.p. and reaches an \aprole of $46.36\%$ when passive mode task is used.

\textbf{Target presence:} Target presence has on \hibiscus{} performance a huge impact, increasing results by about $20$ p.p. Such a variation in performance is due to the difficulty of setting a maximal distance (in the embedding) below which a subject can be considered in interaction with the target. It is well-known that directly thresholding a learnt metric is not trivial. Indeed, metric learning does not constrain absolute distance between samples but only a ranking between them. Target presence task is a way to bypass this issue.

\textbf{Depth of Interaction Net:} The number of blocks used in the Interaction Net has been empirically chosen. A succesion of 8 blocks showed the best result.

\subsubsection{Results on \vcoco{} dataset}

As the proposed method focuses on HOI classification independently of object detection task, it can advantageously use any external object detector at inference time. Indeed, changing the detector does not require to re-train or adapt the network, which is a very interesting property when better object detectors appear in the state of the art.
Consequently, we evaluate our model with two different external object detectors in input: Faster RCNN~\cite{FasterRCNN} with ResNet50 backbone (\textit{\r50}) which is generally used by state-of-the-art methods as a basis to learn interactions, and Faster RCNN with a ResNext101 backbone (\textit{\resnext{}}). For fair comparison, we report $RP_DC_D$ results of Interactiveness \cite{Transferable} approach which corresponds to models trained without extra datasets.

Table \ref{table:vcoco_res} shows the evaluation results of CALIPSO variants compared to state-of-the art methods on \vcoco{} dataset. 
\hibiscus{} reaches the second place behind Interactiveness~\cite{Transferable} but it is computationally far more efficient as we will see in Section~\ref{sec:complexity}. 

Besides, in order to decorrelate object detection task from interaction detection one, we use at inference the perfect object detector and report results in table \ref{table:vcoco_res}. The performance is increased by about $7$ p.p. which shows that the main issue is still the interaction detection (i.e. verb classification and subject-target association).

\begin{table}[h!]
	\begin{center}
		\begin{tabular}{| l | l || c |}
			\hline
			\textbf{Method} &  \textbf{Detector / BB} & \aprolebf (\%)\\
			\hline\hline
			VSRL \cite{VSRL} & \r50 & 31.8  \\
			InteractNet \cite{InteractNet} & \r50 & 40.0 \\
			GPNN \cite{GPNN} & Deform. CNN & 44.0  \\
			iCAN late(early) \cite{ICAN} & \r50 & 44.7 (45.3)  \\
			Xu \cite{Knowledge} & \r50 & 45.9  \\
			\textbf{Interactiveness  \cite{Transferable}} & \textbf{\r50} & \textbf{47.8}\\
			\hline
			Ours & \r50 & 46.36\\
			Ours & \resnext{} & 47.65\\
			Ours & Groundtruth & 54.48\\
			\hline
		\end{tabular}
	\end{center}
	\caption{Evaluation results for \hibiscus{} on \vcoco{} test set compared with state-of-the-art methods. Object detectors or backbones (BB) used are mentioned in the middle column.
	}
	\label{table:vcoco_res}
\end{table}

\subsubsection{Results on HICO-DET dataset}

Since objects in HICO-DET dataset are loosely annotated (many boxes can be assigned to the same object), we adopt the same protocol as \cite{InteractNet} to clean annotation. We use a ResNext101 object detector trained on COCO to detect object and assign the ground truth labels from HICO-DET annotations to the detected objects that highly overlap HICO-DET boxes.

Following the evaluation settings of \cite{HORCNN}, we report the quantitative evaluation of \textit{Full}, \textit{Rare}, and \textit{Non-Rare} interactions on ``\textit{default}'' evaluation setting.
Table \ref{table:hico_res} reports the average precision results of our method on HICO-DET dataset, compared to state-of-the-art HOI detection approaches.
Once again, for fair comparison, we reported methods that only use the dataset without help of additional data, such as linguistic knowledge, from external datasets.

The proposed approach shows competitive results reaching the second place with \resnext{} detector. 

\begin{table}[h!]
	\begin{center}
		\begin{tabular}{|l|ccc|}
			\hline
			&
			\multicolumn{3}{c|}{\textbf{Average Precision (\textit{Default})}} \\
			\hline
			\textbf{Method} & \textbf{Full}  & \textbf{Rare}  & \textbf{Non-Rare}  \\
			\hline\hline
			HO-RCNN \cite{HORCNN}      & 7.81 &	5.37   & 8.54   \\
			InteractNet \cite{InteractNet}  & 9.94 & 7.16 & 10.77    \\
			GPNN \cite{GPNN}       & 13.11 & 9.34 & 14.29        \\
			Xu \cite{Knowledge}      & 14.70 & 13.26 & 15.13        \\
			iCAN  \cite{ICAN}      & 14.84 & 10.45 & 16.15        \\			
			\textbf{Interactiveness  \cite{Transferable} }      & \textbf{17.03} & \textbf{13.42} & \textbf{18.11}        \\
			\hline
			Ours (\r50{}) & 14.31 & 10.43 & 15.46\\
			Ours (\resnext{}) & 14.89 & 11.12 & 16.01\\
			\hline
		\end{tabular}
	\end{center}
	\caption{Evaluation results on HICO-DET test set compared with state-of-the-art methods.}
	\label{table:hico_res}
\end{table}

\subsubsection{Computation Complexity and Time}
\label{sec:complexity}

\begin{figure}[h!]
	\centering
	\includegraphics[width=8.3cm]{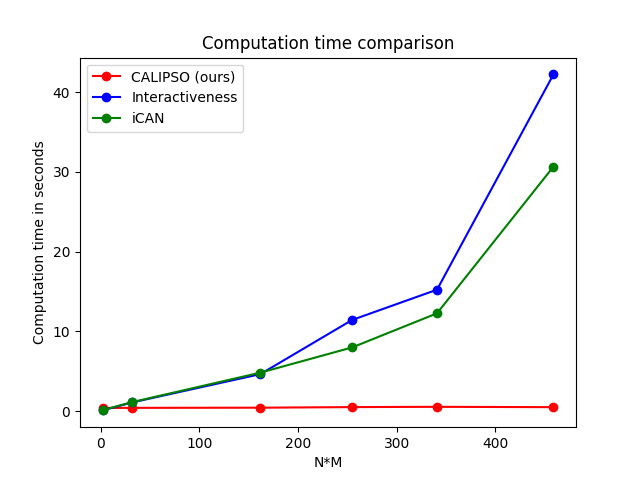}
	\caption{Computation time in seconds for \hibiscus{} (ours), Interactiveness \cite{Transferable} and iCAN \cite{ICAN} for increasing numbers of potential pairs present in images.}
	\label{fig:time}
\end{figure}

Concerning complexity relative to the numbers of people ($N$) and objects ($M$) in the image, notice that \hibiscus{} only does one pass throughout the image with complexity $O(1)$, whereas all other state-of-the-art approaches have a complexity of $O(P)$ with $P$ the number of processed pairs, $T \leq P \leq N \times M$ with $T = |\mathcal{T}|$ the number of ground truth triplets. 
The impact on computation time is shown in Figure \ref{fig:time}: \hibiscus{} runs in constant time (460 ms on NVIDIA Titan X Pascal) independently of the numbers of people and objects in the image. Differently, state-of-the-art methods which provide their codes, Interactiveness \cite{Transferable} and iCAN \cite{ICAN}, have a soaring computation time (e.g., from less than 1 second to more than 40 seconds for Interactiveness).

\section{Conclusion}
\label{sec:conclusion}

In this paper{\let\thefootnote\relax\footnote{{This work was partly supported by Conseil r\'{e}gional d'Ile-de-France.\\Training was performed on Factory-IA, CEA computer facilities.}}}, we proposed a novel interaction detection model, named \hibiscus{}. It estimates all interactions efficiently and simultaneously between all human subjects and object targets by performing a single forward pass throughout the image, regardless of the numbers of objects and interactions in the image.
This constant complexity is achieved thanks to a metric learning strategy that clusters subject and target in interaction, and pushes away all non-interacting objects.
Besides, adding a target presence estimation task as well as an object-centric passive-voice verb estimation for redundancy showed performance improvement.
The proposed method shows competitive results on two widely used datasets, compared to the state of the art, while being much more scalable with the number of interactions in the image.

{\small
\bibliographystyle{ieee}
\bibliography{egpaper_for_review}
}

\end{document}